# Feasibility Study of Urban Flood Mapping Using Traffic Signs for Route Optimization

Bahareh Alizadeh, Diya Li, Zhe Zhang, Amir H. Behzadan
Texas A&M University, USA
abehzadan@tamu.edu

**Abstract.** Water events are the most frequent and costliest climate disasters around the world. In the U.S., an estimated 127 million people who live in coastal areas are at risk of substantial home damage from hurricanes or flooding. In flood emergency management, timely and effective spatial decision-making and intelligent routing depend on flood depth information at a fine spatiotemporal scale. In this paper, crowdsourcing is utilized to collect photos of submerged stop signs, and pair each photo with a pre-flood photo taken at the same location. Each photo pair is then analyzed using deep neural network and image processing to estimate the depth of floodwater in the location of the photo. Generated point-by-point depth data is converted to a flood inundation map and used by an A* search algorithm to determine an optimal flood-free path connecting points of interest. Results provide crucial information to rescue teams and evacuees by enabling effective wayfinding during flooding events.

## 1. Introduction

In recent decades, rapid land development, mass migrations, and deforestation in many parts of the world have overloaded critical infrastructure including road networks and drainage systems especially in at-risk communities and coastal population centers (Sahin & Hall, 1996; Bjorvatn, 2000). This problem is exacerbated by excessive stormwater runoff on impermeable surfaces (e.g., roads, parking lots, driveways, roofs, sidewalks), putting additional strain on the deteriorating drainage systems. The socioeconomic and environmental costs of urban floods can be significant spanning chronic health problems (Du et al., 2010; Paterson et al., 2018), overwhelming insurance claims (Michel-Kerjan, 2010), decreased property values (Bin & Polasky, 2004), lost business income (Browne & Hoyt, 2000), eroded streams and riverbeds (Galay, 1983), and degraded quality of drinking water (Masciopinto et al., 2019).

In the immediate aftermath of a flood event, emergency managers and first responders are tasked with surveying inundated dwellings and neighborhoods, and rescuing those trapped in floodwaters. A key barrier to successful search and rescue (SAR) operation is the limited scope and high variability of field data describing the extent of flood damage and road network vulnerability that could potentially disrupt or prevent timely resource deployment (Keech et al., 2019; Helderop and Grubesic, 2019; Abdullah et al., 2020). Moving floodwater and changing water levels over time necessitates access to (near-) real time floodwater depth information to help people and first responders avoid flooded areas and passages (Liu et al., 2006). In Hurricane Katrina in 2017, for example, emergency responders were frequently querying information about the extent of flood depth to deploy the right type of vehicles for SAR missions and determine the best route for accessing victims (Brecht, 2008). In the absence of such data, people tend to estimate the floodwater depth and level of destruction in their neighbourhoods using social media posts or news stories which can contain outdated data or misinformation (Brecht, 2008; Fan et al., 2020).

In this paper, we conduct a feasibility study with the goal of developing an intelligent spatial decision support system that integrates street-level flood inundation mapping and data-driven routing system using geographic information system (GIS), computer vision, and crowdsourcing. The project aims to support risk-informed spatial decision-making for first responders and communities by providing flood-prone regions with reliable, scalable, and (near-) real time estimation of floodwater depth in the surrounding areas.

## 2. Literature Review

Conventional methods of floodwater depth calculation use sparse data from contact water level gauges, water depth sensors, flood gauges, and water wells (Nair and Rao, 2016; Water Systems Council, 2014; Chetpattananondh et al., 2014; Töyrä et al., 2002; Odli et al., 2016). However, these devices may fail or be washed away in heavy rain (Nair and Rao, 2016). Moreover, water gauges have limited coverage areas (primarily in and around riverine or coastal lands), and need major effort for installation, calibration, and maintenance in flood susceptible locations. Researchers have also used hydrodynamic modeling to estimate flood water depth (Patel et al., 2017; Salimi et al., 2008). However, surface variability and inconsistency (particularly in urban areas) along with the difficulty in differentiating saturated surface soil from standing water in aerial images makes it difficult for these models to yield accurate results. Besides the high cost of sensor installation and operation, a key challenge in floodwater depth analysis in urban places is the low granularity of flood information relative to road and neighborhood data, which makes it extremely difficult to properly overlay road network maps with flood data (Bales and Wagner, 2009; Merwade et al., 2008; Cohen et al., 2018).

In our previous work, we utilized crowdsourcing for large-scale collection of highly granular flood data (Alizadeh Kharazi and Behzadan, 2021). In particular, standardized traffic signs were employed as ubiquitous markers to measure the depth of floodwater in user-contributed photos using artificial intelligence (AI)-based image processing techniques. The motivation behind this approach is the significantly large number of traffic signs that are vastly distributed on the road network in and around residential areas. In the U.S., for example, there are more than 500 types of federally approved traffic signs which have unique shapes and colors, as described in the Manual on Uniform Traffic Control Devices (MUTCD) (FHA, 2004). These signs contain symbols that are recognizable by both humans and computers, e.g., autonomous vehicles use pre-trained models to detect traffic signs on roads (Kurnianggoro et al., 2014). Many such traffic signs are also adopted to a greater degree internationally, thus providing an opportunity for creating a scalable methodological framework for using traffic signs for large-area flood inundation mapping. In this paper, we built upon our past work by generating practical movement plans for evacuees and first responders based on the crowdsourced data through implementing a routing optimization model on a street-level flood inundation map.

The routing problem is one of the most studied combinatorial optimization problems, first mentioned in 1959 as *truck dispatching problem* to determine an optimal route for a fleet of gasoline delivery trucks between a terminal and a number of service stations (Dantzig and Ramser, 1959). One classic variant of this problem is routing in the presence of obstacles (Golden et al., 2008), which is directly applicable to scenarios where people, rescue teams, or other resources need to evacuate disaster-affected areas while avoiding hazardous encounters (e.g., debris fields, flooded areas, blocked roads). In computational geometry, the *watchman route problem* (Chin and Ntafos, 1986) attempts to solve this scenario by computing the shortest route that a watchman should take to guard a particular area with obstacles. Previous research has used polynomial time algorithms to find the shortest route given an area on a map with preset conditions (Carlsson et al., 1999; Tan 2001; Chin and Ntafos, 1986). In graph theory, the same can be modeled as an optimization problem where the goal is to find the shortest path between a subset of nodes. Dijkstra's algorithm is one of the widely recognized solutions to this optimization problem. For example, Li and Klette (2006) used a rubber-band algorithm to find the shortest path between two points in a graph with $O(n \ log \ n)$ time complexity. Other researchers have investigated routing problems in real world cases such as flood events and proposed various algorithms (Wang and Zlatanova, 2013; Kapoor et al., 2007; Golden et al., 2008; Lu et al., 2003). For instance, Lu et al. (2003) designed a capacity-constrained routing algorithm with heuristic methods that incorporated evacuation time to perform route planning with avoidance. More recently, several commercial applications and open-source solutions have been developed to provide convenient routing services. Among others, examples include Nedkov and Zlatanova (2011) who used Google Directions API to extend web direction for routing with avoidance, as well as Engelmann et al. (2020) who used GraphHopper (Karich and Schröder, 2014) to create a route planning method that minimizes the emission of harmful gases from vehicles. However, existing decision support systems for emergency management lack the ability to integrate street-level flood information, risk-informed routing system, and spatial decision-making capabilities, which could weaken the efficiency of the SAR operations. As described in this paper, we incorporate floodwater depth information as an additional constraint into the routing problem to produce practical movement plans for evacuees and first responders.

## 3. Methodology

**Line detection for pole length calculation.** We use stop signs as standardized measurement benchmarks and estimate the depth of the flood by comparing the length of the visible portion of the pole (on which the stop sign is mounted) in pre- and post-flood photos taken from the same location. As shown in Figure 1, Mask Regional Convolutional Neural Network (in short, Mask R-CNN), an object detection and instance segmentation model (He et al., 2017), is used to detect stop signs in paired stop sign photos of the same location prior and after the flood event. After the stop sign is detected, two image processing techniques, namely Canny edge detector (Ogawa et al., 2010; Rong et al., 2014) and probabilistic Hough transform (Zhu and Brilakis, 2009), are used to detect the sign pole in each photo and estimate their lengths. Using this technique, first, all possible edges in each photo are explored. Selected edge candidates are then merged to reconstruct and measure the length of the straight line that is likely to represent the sign pole. The depth of floodwater is subsequently calculated as the difference in pole lengths in paired pre- and post-flood photos. A detailed description of this step is beyond the scope of this paper and can be found in Alizadeh Kharazi and Behzadan (2021).

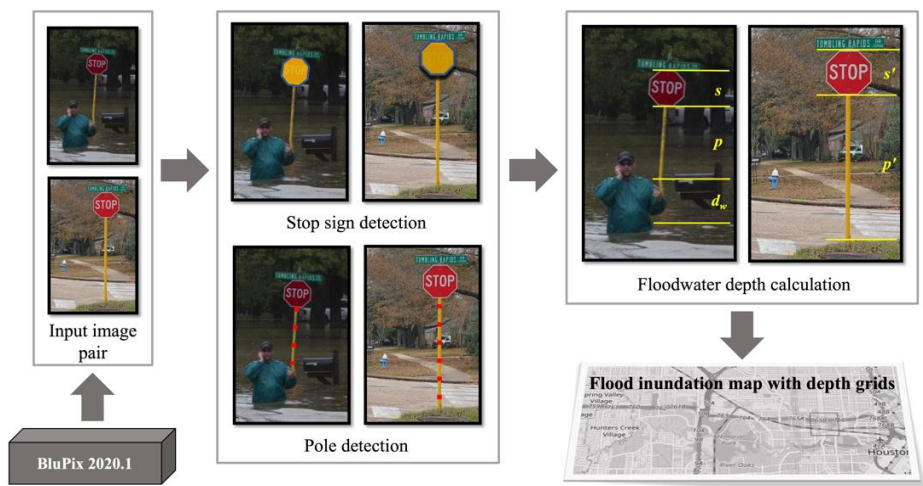

Figure 1: Framework for Estimation of Floodwater Depth by Visual Analysis of Paired Stop Sign Flood Photos. (base post-flood photo: courtesy of Erich Schlegel/Getty Images)

**Simulate flooded areas using GIS and Stop Sign detection results.** Volunteered Geographic Information (VGI) has been utilized in flood studies (Goodchild 2007; Huang et al., 2018). Huang et al. (2018) used an inverse distance weighted height filter to build a probability index distribution (PID) layer from the high-resolution digital elevation model data. Inspired by that work, we use a distance-decay function along with isohypse information to transform point-by-point floodwater depth data into area-wide flood inundation maps. Several forms of this function are widely used to describe systematic spatial variations where spatial information has the tendency to vanish with distance (Haining, 2001). In this research, we create a flooding confidence area around detected floodwater depth points to simulate flooded regions. In the future, this approach can be compared and validated using hydrological-based modeling (Liu et al., 2006) to improve the accuracy of flood inundation mapping. Suppose an estimated flooded area $A$ that is defined by a discrete point grid $\{X_1, X_2, \ldots, X_i, \ldots, X_j\}$. The Gaussian buffering function shown in Equation 1 is applied to approximate the depth of floodwater at point $X_j$ in area A. In this Equation, $X_0$ is the detected floodwater depth at the center point of area $A$, $I_0$ is the elevation at the center point of area $A$, $I_j$ is the elevation in point $j$, $d_j$ is the geographic distance between the center point of area $A$ and $X_j$, and $b$ is a fixed bandwidth for the Gaussian function. As the distance $d_j$ varies around area $A$, the estimated floodwater depth also changes with distance-decay and isohypse information. This approach is commonly used in GIS research such as social-media flood mapping (Huang et al., 2018) and distance-decay weight regression model (Gutiérrez et al., 2011).

$$X_j = X_0 \, exp\, [-1/2(d_j/b)^2] + (I_0 - I_j) \tag{1}$$

**Description of the routing problem.** Given the flood inundation information, the routing problem from an origin to a destination point can be modeled as a multi-stage decision process, where each decision stage includes the location of the current decision point as well as the time needed to complete the remainder of the process. The designed optimization-based algorithm proposes a routing solution that avoids flood inundated areas and supports SAR operations during a flood event. The routing problem is further modified by including several decision objectives, and transforming the otherwise single-objective optimization into a multi-objective decision process. From the taxonomy of navigation for emergency response (Wang and Zlatanova, 2013), this problem can be defined using $X = <X_1, X_2, X_3, X_4, \ldots>$, where $X_i$ denotes an environment factor, and contains the quantity (one or many), and the type (e.g., destination, responder object, obstacle) of that factor. For example, for a person whose goal is to go back home while avoiding flooded roads, the navigation route can be defined as $<\{one\ moving\ object\}, \{one\ static\ destination\}, \{one\ static\ obstacles\}>$. Since the traditional Dijkstra algorithms may not work well for $X$ that contains obstacle factors, we propose to use the A* search algorithm (Russell and Norvig, 2002; Lerner et al., 2009). For this purpose, the following concepts are adopted before we formalize each iteration of the A* search algorithm:

1. Search area: Given a prepared base map that contains spatial entities, each entity is represented as a graph node in the search area.

2. Open vs. closed list: All nearest nodes waiting to be searched are stored in an open list; and those already searched are stored in a closed list.

3. Path sorting: To define the direction of the next movement, we use a path sorting function expressed by Equation 2.

$$F(n) = G + H \tag{2}$$

In this Equation, *H* denotes a heuristic function, and *G* is the moving cost from the initial location to the next node in the open list. The heuristic function takes the Manhattan distance to calculate the cost of moving from each of the candidate next nodes in the open list to the final destination node. Figure 2 presents the pseudo algorithm for the A* search. A demonstration of how this algorithm is used in a flooded area (routing with obstacles) is shown in Figure 3. In this Figure, different shades of blue represent different floodwater depth values, diamonds stand for start and destination points, and orange pixels mark the simulated routing path. The cost value is displayed in each searched pixel. This information is used as threshold conditions to check whether a vehicle can pass through a particular area.

```
Initialize open_list and close_list, start with initial point;
Add start node s to open_list;  F(s) = 0 (smallest);
If open_list is not Null, select node n with smallest F(n):
  If n is destination node:
    Find parent node from destination node to start node then return;
  If n is not destination node:
    Move n from open_list to close_list;
    Traverse eight nearest nodes of n:
      If nearest node m in close_list: continue;
      If nearest node not in open_list:
        Set parent node n for m; Update cost function; add m to open_list
Connect parent node from initial node and generate path;
```

Figure 2: Pseudo Algorithm for A* Search.

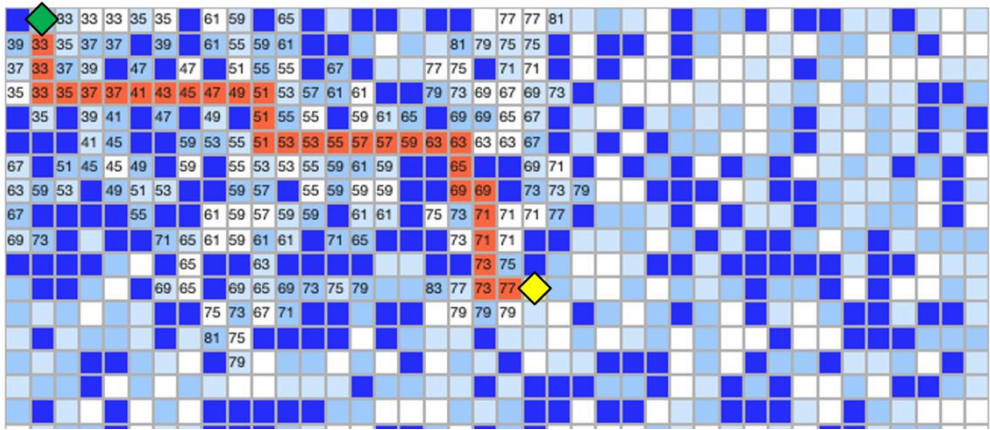

Figure 3: Sample Output of A* Search Algorithm with Obstacles.

## 4. Proof-of-Concept Experiment

As shown in Figure 4, for the flood scenario presented in this paper, six paired photos from the 2017 Hurricane Harvey in Houston, Texas, taken approximately on the same date in the month of September, are selected from BluPix v.2020.1, a crowdsourcing platform developed in this research to collect user-contributed photos of flooded stop signs.

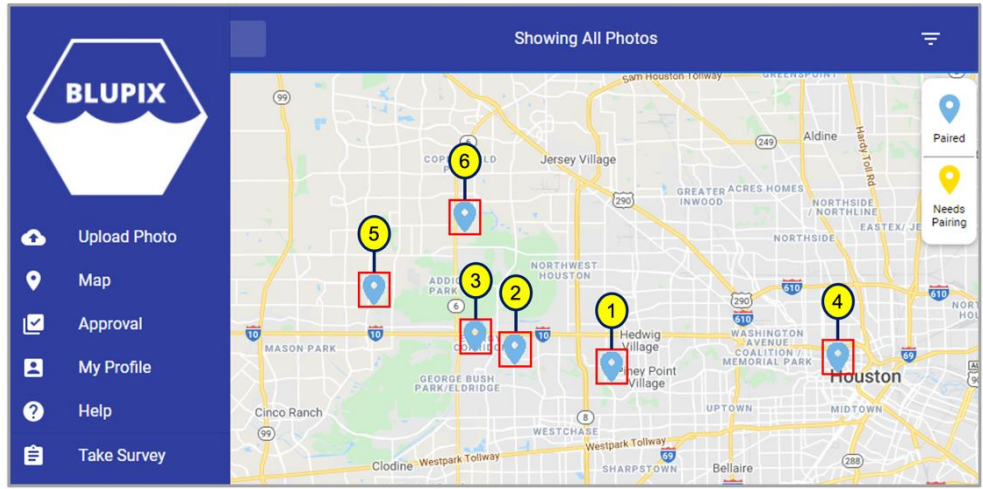

Figure 4:   Locations of Selected Paired Flood Photos in Houston, TX after Hurricane Harvey (2017).

Table 1 shows a summary of floodwater depth calculations applied to pre- and post-flood photos. As shown in this Table, the root mean square error (RMSE) of the flood depth estimation model on the six pairs of pre- and post-flood photos is 4.69 inches, and the average processing time for floodwater depth calculation is 11.6 seconds.

Table 1:   Performance of floodwater depth estimation on paired pre- and post-flood photos.

| Calculation | Metric | Pre-flood photos ($n = 6$) | Post-flood photos ($n = 6$) |
| --- | --- | --- | --- |
| Stop sign detection | Intersection over union (IOU) % | 96.73 | 95.24 |
| | Precision % | 100 | 100 |
| | Recall % | 100 | 100 |
| | Average precision (AP) % | 100 | 100 |
| | Average processing time (s) | 1.08 | 1.26 |
| Pole detection | RMSE (in.) | 2.64 | 5.80 |
| | Average processing time (s) | 0.13 | 0.10 |
| Flood depth estimation | Average total processing time (s) | 2.57 | |
| | RMSE (in.) | 4.69 | |

Floodwater depth estimates are subsequently used to generate a flood inundation map with depth grids. Figure 5 demonstrates the application of A* search algorithm to calculate the shortest flood-free route. In this example, each of the previously selected six paired points is taken as the central point of a flooded area. To implement the distance-decay function (Equation 1), elevation data is queried from Google Elevation API. The Graphhopper library (Karich and Schröder, 2014) is used for route search, and Openrouteservice is utilized to overlay the base map with generated flooded areas. The basic spatial information for building the base map is taken from OpenStreetMap (Planet OpenStreetMap, 2021).

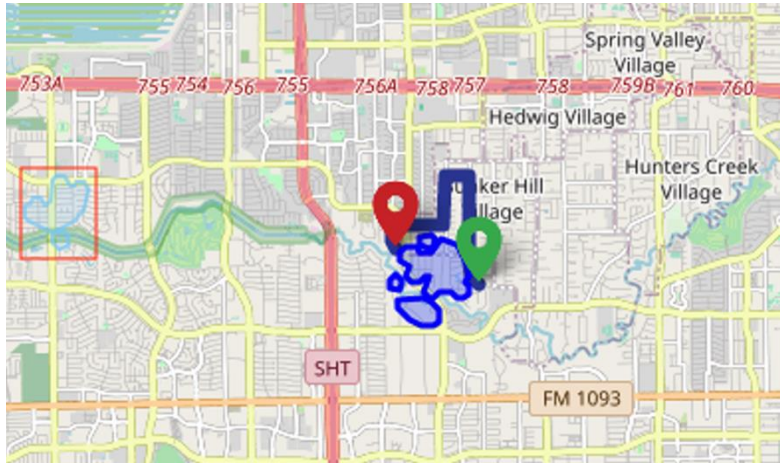

Figure 5:   Illustration of the routing algorithm using buffered points that represent estimate floodwater depths collected from inundated areas.

## 5.  Summary and Conclusion

Flood is the most common type of climate disaster in the U.S. and around the world. An impediment to timely SAR and routing of resources during flood events is the lack of street-level floodwater depth information. Since water levels change over time, it is necessary to have access to (near-) real time floodwater depth information to help people and first responders avoid flooded areas and passages. This paper proposed the use of standardized traffic signs as ubiquitous markers for measuring the depth of floodwater in user-contributed street photos. We used Mask R-CNN, a deep neural network, to detect stop signs in photos, and applied two image processing techniques (Canny edge detector and probabilistic Hough transform) to determine the length of the sign pole in pre- and post-flood photos. Floodwater depth was then estimated as the difference between pole lengths of the same stop sign in paired pre- and post-flood photos. We achieved an RMSE of 4.69 inches in estimating the floodwater depth for a set of six paired flood photos taken in Houston, TX after the 2017 Hurricane Harvey.

Next, distance-decay function was implemented to transform point-by-point floodwater depth data into area-wide flood inundation maps. The generated map was used to develop a risk-informed routing system based upon the A* search algorithm to calculate the shortest flood-free route between points of interest (i.e., intelligent wayfinding). In the current implementation of the route optimization algorithm, all flooded areas are avoided. However, one can set a threshold value and customize their search based on the vehicle type used to navigate in a flooded area. Clearly, increased public awareness and improved user experience can help gather a large number of floodwater depth points and improve the reliability of the algorithm. In the meantime, when only limited data points are available for flood mapping, additional flood depth data can be generated using advanced hydrological models. The minimum number of flood depth data needed for flood mapping may also depend on the type, topography, and other characteristics of the flooded surface. In a flat area, for example, fewer data points are generally sufficient to generate accurate flood maps. In contrast, more points may be needed in rugged areas or where surface characteristics and shape change abruptly. In this paper, using a Gaussian distance-decay function with each pole independently allowed us to relax the minimum number requirement when generating flood maps with sparse data.

The developed methods in this paper are sought to interface with other sources of spatial information (e.g., high-resolution point cloud terrain), leading to further improvement of flood mapping and wayfinding. Moreover, paired photos are stored with time and location information, allowing a host of spatiotemporal analyses of past flood events. All in all, the designed route optimization will provide crucial information to rescue teams and evacuees and enable effective wayfinding during flooding events. In the long term, the generalizability and robustness of the designed platform will be rigorously evaluated as more user-contributed photos are collected and paired on the BluPix crowdsourcing application.

**Acknowledgments**

This study is funded by award #NA18OAR4170088 from the National Oceanic and Atmospheric Administration (NOAA), U.S. Department of Commerce. We are thankful to Dr. Courtney Thompson and Dr. Michelle Meyer (project collaborators at Texas A&M University) for their supporting role in this project. Additionally, we would like to thank Mr. Nathan Young (undergraduate student at Texas A&M University) for his assistance in data collection. Any opinions, findings, conclusions, and recommendations expressed in this paper are those of the authors and do not necessarily represent the views of the NOAA, Department of Commerce, or the individuals named above.

**References**

Abdullah, M., Suliman, M.S., Daud, M.S.M., Hamid, Z.J.M.H., Noor, M.R.M. Ngadiman, N.I. (2019). Humanitarian logistic relief team challenges during flood. In: International Research Conference and Innovation Exhibition, 2019, Johor Bahru, Malaysia.

Alizadeh Kharazi, B., & Behzadan, A. H. (2021). Flood depth mapping in street photos with image processing and deep neural networks. Computers, Environment and Urban Systems, 88, pp.101628.

Bales, J.D., Wagner, C.R. (2009). Sources of uncertainty in flood inundation maps, J. Flood Risk Management, 2(2), pp.139–147.

Bin, O., Polasky, S. (2004). Effects of flood hazards on property values: evidence before and after Hurricane Floyd, J. Land Economics, 80(4), pp.490–500.

Bjorvatn, K. (2000). Urban infrastructure and industrialization, J. Urban Economics, 48(2), pp.205–218.

Brecht, H. (2008). The application of geo-technologies after Hurricane Katrina. In: Nayak, S., Zlatanova, S. (Eds.). Remote Sensing and GIS Technologies for Monitoring and Prediction of Disasters, Berlin: Springer-Verlag, pp.281–304.

Browne, M.J., Hoyt, R.E. (2000). The demand for flood insurance: empirical evidence, J. Risk and Uncertainty, 20(3), 291–306.

Carlsson, S., Jonsson, H., Nilsson, B.J. (1999). Finding the shortest watchman route in a simple polygon, J. Discrete and Computational Geometry, 22(3), pp.377–402.

Chetpattananondh, K., Tapoanoi, T., Phukpattaranont, P., Jindapetch, N. (2014). A self-calibration water level measurement using an interdigital capacitive sensor, J. Sensors and Actuators A: Physical, 209, pp.175–182.

Chin, W.P., Ntafos, S. (1986). Optimum watchman routes. In: Second Annual Symposium on Computational Geometry, 1986, New York, NY.

Cohen, S., Brakenridge, G.R., Kettner, A., Bates, B., Nelson, J., McDonald, R., Huang, Y.F., Munasinghe, D., Zhang, J. (2018), Estimating floodwater depths from flood inundation maps and topography, JAWRA J. of the American Water Resources Association, 54(4), pp.847–858.

Dantzig, G.B., Ramser, J.H. (1959). The truck dispatching problem, J. Management Science, 6(1), pp.80–91.

Du, W., Fitzgerald, G.J., Clark, M., Hou, X.Y. (2010). Health impacts of floods, J. Prehospital and Disaster Medicine, 25(3), pp.265–272.

Engelmann, M., Schulze, P., Wittmann, J. (2020). Emission-based routing using the GraphHopper API and OpenStreetMap. In: Schaldach, R., Simon, K.H., Weismüller, J., Wohlgemuth, V. (Eds.). Advances and New Trends in Environmental Informatics, Progress in IS. Springer, Cham, pp.91–104.

Fan, C., Esparza, M., Dargin, J., Wu, F., Oztekin, B., Mostafavi, A. (2020). Spatial biases in crowdsourced data: Social media content attention concentrates on populous areas in disasters. J. Computers, Environment and Urban Systems, 83, 101514.

FHA, (2004). Manual of Uniform Traffic Control Devices (MUTCD): Standard Highway Signs. Washington: Federal Highway Administration.

Galay, V.J. (1983). Causes of riverbed degradation, J. Water Resources Research, 19(5), pp.1057–1090.

Golden, B.L., Raghavan, S., Wasil, E.A. (2008). The Vehicle Routing Problem: Latest Advances and New Challenges (Vol. 43). Berlin: Springer-Verlag.

Goodchild, M. F. (2007). Citizens as sensors: The world of volunteered geography, GeoJournal, 69(4), 211–221.

Gutiérrez, J., Cardozo, O.D., García-Palomares, J.C. (2011). Transit ridership forecasting at station level: An approach based on distance-decay weighted regression, J. Transport Geography, 19(6), pp.1081–1092.

Haining, R.P. (2001). Spatial Sampling. In: Smelser, N.J., Baltes, P.B. (Eds.). International Encyclopedia of the Social and Behavioral Sciences. Oxford: Pergamon.

He, K., Gkioxari, G., Dollár, P., Girshick, R. (2017). Mask R-CNN. In: IEEE International Conference on Computer Vision (ICCV), 2017, Venice, Italy.

Helderop, E., Grubesic, T.H. (2019). Streets, storm surge, and the frailty of urban transport systems: A grid-based approach for identifying informal street network connections to facilitate mobility. Transportation Research Part D: Transport and Environment, 77, 337–351.

Huang, X., Wang, C., Li, Z. (2018). A near real-time flood-mapping approach by integrating social media and post-event satellite imagery, Annals of GIS, 24(2), 113–123.

Kapoor, S., Maheshwari, S.N., Mitchell, J.S. (1997). An efficient algorithm for Euclidean shortest paths among polygonal obstacles in the plane, J. Discrete and Computational Geometry, 18(4), pp.377–383.

Karich, P., Schröder, S. (2014). Graphhopper, http://www.graphhopper.com, accessed March 2021.

Keech, J.J., Smith, S.R., Peden, A.E., Hagger, M.S. Hamilton, K. (2019). The lived experience of rescuing people who have driven into floodwater: Understanding challenges and identifying areas for providing support, Health Promotion J. of Australia, 30(2), pp.252–257.

Kurnianggoro, L., Hariyono, J., Jo, K.H. (2014). Traffic sign recognition system for autonomous vehicle using cascade SVM classifier. In: 40th Annual Conference of the IEEE Industrial Electronics Society, 2014, Dallas, TX.

Lerner, J., Wagner, D., Zweig, K. (2009). Algorithmics of Large and Complex Networks: Design, Analysis, and Simulation (Vol. 5515). Berlin: Springer-Verlag.

Li, F., Klette, R. (2006). Finding the shortest path between two points in a simple polygon by applying a rubberband algorithm. In: Pacific-Rim Symposium on Image and Video Technology, 2006, Hsinchu, Taiwan.

Liu, Y., Hatayama, M., Okada, N. (2006). Development of an adaptive evacuation route algorithm under flood disaster. Annuals of Disaster Prevention Research Institute, Kyoto University, 49, 189–195.

Lu, Q., Huang, Y., Shekhar, S. (2003). Evacuation planning: a capacity constrained routing approach. In: International Conference on Intelligence and Security Informatics, 2003, Tucson, AZ.

Masciopinto, C., De Giglio, O., Scrascia, M., Fortunato, F., La Rosa, G., Suffredini, E., Pazzani, C., Prato, R., Montagna, M.T. (2019). Human health risk assessment for the occurrence of enteric viruses in drinking water from wells: Role of flood runoff injections, J. Science of the Total Environment, 666, pp.559–571.

Merwade, V., Olivera, F., Arabi, M., Edleman, S. (2008). Uncertainty in flood inundation mapping: current issues and future directions, J. Hydrologic Engineering, 13(7), 608–620.

Michel-Kerjan, E.O. (2010). Catastrophe economics: the national flood insurance program, J. Economic Perspectives, 24(4), pp.165–186.

Nair, B.B., Rao, S. (2016). Flood water depth estimation: A survey. In: IEEE International Conference on Computational Intelligence and Computing Research (ICCIC), 2016, Tamil Nadu, India.

Nedkov, S., Zlatanova, S. (2011). Enabling obstacle avoidance for Google maps' navigation service. In: GeoInformation for Disaster Management, 2011, Antalya, Turkey.

Odli, Z.S.M., Izhar, T.N.T., Razak, A.R.A., Yusuf, S.Y., Zakarya, I.A., Saad, F.N.M., Nor, M.Z.M. (2016). Development of portable water level sensor for flood management system, J. Engineering and Applied Sciences, 11(8), pp.5352–5357.

Ogawa, K., Ito, Y., Nakano, K. (2010). Efficient Canny edge detection using a GPU. In: IEEE First International Conference on Networking and Computing, 2010, Higashi, Japan.

Planet OpenStreetMap. (2021). Planet dump [Data file from June 4th. 2017 dumps], https://planet.openstreetmap.org, accessed May 2021.

Patel, D.P., Ramirez, J.A., Srivastava, P.K., Bray, M., Han, D. (2017). Assessment of flood inundation mapping of Surat city by coupled 1D/2D hydrodynamic modeling: A case application of the new HEC-RAS 5, J. Natural Hazards, 89(1), pp.93–130.

Paterson, D.L., Wright, H., Harris, P.N. (2018). Health risks of flood disasters, J. Clinical Infectious Diseases, 67(9), pp.1450–1454.

Rong, W., Li, Z., Zhang, W., Sun, L. (2014). An improved Canny edge detection algorithm. In: IEEE International Conference on Mechatronics and Automation, 2014, Tianjin, China.

Russell, S., Norvig, P. (2002). Artificial Intelligence: A Modern Approach. Upper Saddle River: Prentice Hall.

Sahin, V., Hall, M.J. (1996). The effects of afforestation and deforestation on water yields, J. Hydrology, 178(1–4), pp.293–309.

Salimi, S., Ghanbarpour, M.R., Solaimani, K., Ahmadi, M.Z. (2008). Floodplain mapping using hydraulic simulation model in GIS, J. Applied Sciences, 8, pp.660–665.

Tan, X. (2001). Fast computation of shortest watchman routes in simple polygons, Information Processing Letters, 77(1), pp.27–33.

Töyrä, J., Pietroniro, A., Martz, L.W., Prowse, T.D. (2002). A multi-sensor approach to wetland flood monitoring, J. Hydrological Processes, 16(8), pp.1569–1581.

Wang, Z., Zlatanova, S. (2013). Taxonomy of navigation for first responders. In: Krisp, J.M. (Ed.). Progress in Location-Based Services, Berlin: Springer-Verlag, pp.297–315.

Water Systems Council (2020). Well Owner's Manual, https://www.watersystemscouncil.org/download/3430/, accessed February 2021.

Zhu, Z., Brilakis, I. (2009). Comparison of optical sensor-based spatial data collection techniques for civil infrastructure modelling, J. Computing in Civil Engineering, 23(3), 170–177.